\documentclass[11pt]{article}
\usepackage{eacl2014}
\usepackage{times}
\usepackage{url}
\usepackage{latexsym}
\usepackage{amsfonts,amstext,amsmath,amssymb}
\usepackage{breqn}

\usepackage{graphicx}
\usepackage{caption}
\usepackage{subcaption}

\title{Word Embeddings through Hellinger PCA}

\author{R\'emi Lebret\\
	    Idiap Research Institute \\
	    Rue Marconi 19, CP 592 \\
	    1920 Martigny, Switzerland \\
	    {\tt remi@lebret.ch}
	  \And
	  Ronan Collobert\\
  	Idiap Research Institute \\
	Rue Marconi 19, CP 592 \\
	1920 Martigny, Switzerland \\
 	 {\tt ronan@collobert.com}}

\date{}

\begin{document}
\maketitle
\begin{abstract}
Word embeddings resulting from neural language models have been shown to be a great asset for a large variety of NLP tasks.
However, such architecture might be difficult and time-consuming to train.
Instead, we propose to drastically simplify the word embeddings computation through a Hellinger PCA of the word co-occurence matrix.
We compare those new word embeddings with some well-known embeddings on named entity recognition and movie review tasks and show that we can reach similar or even better performance.
Although deep learning is not really necessary for generating good word embeddings, we show that it can provide an easy way to adapt embeddings to specific tasks.
\end{abstract}

\section{Introduction}

Building word embeddings has always generated much interest for linguists.
Popular approaches such as Brown clustering algorithm~\cite{brown:1992} have
been used with success in a wide variety of NLP
tasks~\cite{schutze:1995,koo:2008,ratinov:2009}.  Those word embeddings are
often seen as a low dimensional-vector space where the dimensions
are features potentially describing syntactic or semantic
properties. Recently, distributed approaches based on neural network
language models (NNLM) have revived the field of learning word embeddings
\cite{collobert:2008,huang:2009,TurianRaBe2010,collobert:2011b}.
However, a neural network architecture can be hard to train. Finding the
right parameters to tune the model is often a challenging task and the training phase is in general computationally expensive.

This paper aims to show that such good word embeddings can be obtained
using simple (mostly linear) operations.  We show that similar word embeddings can
be computed using the word co-occurrence statistics and a well-known
dimensionality reduction operation such as Principal Component Analysis
(PCA).  We then compare our embeddings with the CW \cite{collobert:2008}, Turian \cite{TurianRaBe2010}, HLBL \cite{citeulike:4192404} embeddings, which come from deep architectures and the LR-MVL \cite{dhillon11multiviewcca}
embeddings, which also come from a spectral method on several NLP tasks.

We claim that, assuming an appropriate metric, a simple spectral method as PCA can generate word embeddings as good as with deep-learning architectures.
On the other hand, deep-learning architectures have shown their potential in several supervised NLP tasks, by using these word embeddings.
As they are usually generated over large corpora of unlabeled data, words are represented in a generic manner.
Having generic embeddings, good performance can be achieved on NLP tasks where the syntactic aspect is dominant such as Part-Of-Speech, chunking and NER \cite{TurianRaBe2010,collobert:2011b,dhillon11multiviewcca}.
For supervised tasks relying more on the semantic aspect as sentiment classification, it is usually helpful to adapt the existing embeddings to improve performance \cite{labutov13}.
We show in this paper that such embedding specialization can be easily done via neural network architectures and that helps to increase general performance.

\section{Related Work}

As 80\% of the meaning of English text comes from word choice and the remaining 20\% comes from word order \cite{landauer02}, it seems quite important to leverage word order to capture all the semantic information.
Connectionist approaches have therefore been proposed to develop distributed representations which encode the structural relationships between words \cite{hinton:learndistrep,Pollack90recursivedistributed,DBLP:journals/ml/Elman91}.
More recently, a neural network language model was proposed in \newcite{Bengio:2003:NPL:944919.944966} where word vector representations are simultaneously learned along with a statistical language model. This architecture inspired other authors: \newcite{collobert:2008} designed a neural language model which eliminates the linear dependency on vocabulary size, \newcite{citeulike:4192404} proposed a hierarchical linear neural model, \newcite{Mikolov2010} investigated a recurrent neural network architecture for language modeling.
Such architectures being trained over large corpora of unlabeled text with the aim to predict correct scores end up learning the co-occurence statistics.

Linguists assumed long ago that words occurring in similar contexts tend to have similar meanings \cite{Wittgenstein1953}.
Using the word co-occurrence statistics is thus a natural choice to embed similar words into a common vector space \cite{Turney2010}.
Common approaches calculate the frequencies, apply some transformations (tf-idf, PPMI), reduce the dimensionality and calculate the similarities \cite{Lowe2001}.
Considering a fixed-sized word vocabulary $\mathcal{D}$ and a set of words $\mathcal{W}$ to embed, the co-occurence matrix $C$ is of size $|\mathcal{W}| \times |\mathcal{D}|$. $C$ is then vocabulary size-dependent.
One can apply a dimensionality reduction operation to $C$ leading to $\bar{C}\in \mathbb{R}^{|\mathcal{W}| \times d}$, where $d \ll |\mathcal{D}|$.
Dimensionality reduction techniques such as Singular Valued Decomposition (SVD) are widely used (e.g. LSA \cite{Landauer1997}, ICA \cite{Vayrynen04STeP2004}).
However, word co-occurence statistics are discrete distributions.
An information theory measure such as the Hellinger distance seems to be more appropriate than the Euclidean distance over a discrete distribution space.
In this paper we will compare the Hellinger PCA against the classical Euclidean PCA and the Low Rank Multi-View Learning (LR-MVL) method, which is another spectral method based on Canonical Correlation Analysis (CCA) to learn word embeddings \cite{dhillon11multiviewcca}.

It has been shown that using word embeddings as features helps to improve general performance on many NLP tasks \cite{TurianRaBe2010}.
However these embeddings can be too generic to perform well on other tasks such as sentiment classification.
For such task, word embeddings must capture the sentiment information.
\newcite{MaasDPHNP11} proposed a model for jointly capturing semantic and sentiment components of words into vector spaces.
More recently, \newcite{labutov13} presented a method which takes existing embeddings and, by using some labeled data, re-embed them in the same space.
They showed that these new embeddings can be better predictors in a supervised task.
In this paper, we consider word embedding-based linear and non-linear models for two NLP supervised tasks: Named Entity Recognition and IMDB movie review.
We analyze the effect of fine-tuning existing embeddings over each task of interest.

\section{Spectral Method for Word Embeddings}
\label{method}

A NNLM learns which words among the vocabulary are likely to appear after a
given sequence of words. More formally, it learns the next word
probability distribution. Instead, simply counting words on a large corpus
of unlabeled text can be performed to retrieve those word distributions and
to represent words~\cite{Turney2010}.

\subsection{Word co-occurence statistics}

"You shall know a word by the company it keeps" \cite{firth57synopsis}.  It
is a natural choice to use the word co-occurence statistics to acquire
representations of word meanings.  Raw word co-occurence frequencies are
computed by counting the number of times each context word $w \in
\mathcal{D}$ occurs after a sequence of words $T$:
\begin{equation}
p(w|T)=\frac{p(w,T)}{p(T)}=\frac{n(w,T)}{\sum_{w}{n(w,T)}}\,,
\end{equation}
where $n(w,T)$ is the number of times each context word $w$ occurs after the sequence $T$. The size of $T$ can go from 1 to $t$ words.
The next word probability distribution $p$ for each word or sequence of words is thus obtained. It is a multinomial distribution of
$|\mathcal{D}|$ classes (words).  A co-occurence matrix of size $N \times
|\mathcal{D}|$ is finally built by computing those frequencies over all the
$N$ possible sequences of words.

\subsection{Hellinger distance}
Similarities between words can be derived by computing a distance between
their corresponding word distributions. Several distances (or metrics) over
discrete distributions exist, such as the Bhattacharyya distance, the
Hellinger distance or Kullback-Leibler divergence. We chose here the
Hellinger distance for its simplicity and symmetry property (as it is a
true distance). Considering two discrete probability distributions $P=(p_1,\ldots,p_k)$
and $Q=(q_1,\ldots,q_k)$, the Hellinger distance is formally defined as:
\begin{equation}
H(P,Q)=\frac{1}{\sqrt{2}}\sqrt{\sum_{i=1}^{k}{(\sqrt{p_i}-\sqrt{q_i})^2}}\,,
\end{equation}
which is directly related to the Euclidean norm of the difference of the square root vectors:
\begin{equation}
H(P,Q)=\frac{1}{\sqrt{2}}\lVert\sqrt{P}-\sqrt{Q}\rVert_2\,.
\end{equation}
Note that it makes more sense to take the Hellinger distance rather than
the Euclidean distance for comparing discrete distributions, as $P$ and $Q$
are unit vectors according to the Hellinger distance ($\sqrt{P}$ and
$\sqrt{Q}$ are units vector according to the $\ell_2$ norm).
\subsection{Dimensionality Reduction}
As discrete distributions are vocabulary size-dependent, using
directly the distribution as a word embedding is not really tractable for
large vocabulary. We propose to perform a principal component analysis
(PCA) of the word co-occurence probability matrix to
represent words in a lower dimensional space while minimizing the
reconstruction error according to the Hellinger distance.

\section{Architectures for NLP tasks}

Traditional NLP approaches extract from documents a rich set of hand-designed features which are then fed to a standard classification algorithm.
The choice of features is a task-specific empirical process.
In contrast, we want to pre-process our features as little as possible.
In that respect, a multilayer neural network architecture seems appropriate as it can be trained in an end-to-end fashion on the task of interest.

\subsection{Sentence-level Approach}
\label{sent}

The sentence-level approach aims at tagging with a label each word in a given sentence.
Embeddings of each word in a sentence are fed to linear and non-linear classification models followed by a CRF-type sentence tag inference.
We chose here neural networks as classifiers.

\paragraph{Sliding window}
Context is crucial to characterize word meanings. We thus consider $n$ context words around each word $x_t$ to be tagged, leading to a window of $N=(2n+1)$ words $[x]^t=(x_{t-n},\ldots, x_t, \ldots, x_{t+n})$.
As each word is embedded into a $d_{wrd}$-dimensional vector, it results a $d_{wrd} \times N$ vector representing a window of $N$ words, which aims at characterizing the middle word $x_t$ in this window.
Given a complete sentence of $T$ words, we can obtain for each word a context-dependent representation by sliding over all the possible windows in the sentence.
A same linear transformation is then applied on each window for each word to tag:
\begin{equation}
g([x]^t)=W[x]^t+b\,,
\end{equation}
where $W \in \mathbb{R}^{M \times d_{wrd}N}$ and $b \in \mathbb{R}^{M}$ are the parameters, with $M$ the number of classes.
Alternatively, a one hidden layer non-linear network can be considered:
\begin{equation}
g([x]^t)=Wh(U[x]^t)+b\,,
\end{equation}
where $U\in \mathbb{R}^{n_{hu} \times d_{wrd}N}$, with $n_{hu}$ the number of hidden units and $h(.)$ a transfer function.
\paragraph{CRF-type inference}
There exists strong dependencies between tags in a sentence: some tags cannot follow other tags.
To take the sentence structure into account, we want to encourage valid paths of tags during training, while discouraging all other paths.
Considering the matrix of scores outputs by the network, we train a simple conditional random field (CRF).
At inference time, given a sentence to tag, the best path which minimizes the sentence score is inferred with the Viterbi algorithm.
More formally, we denote $\theta$ all the trainable parameters of the network and $f_\theta([x]_1^T)$ the matrix of scores. The element $[f_\theta]_{i,t}$ of the matrix is the score output by the network for the sentence $[x]_1^T$ and the $i^{th}$ tag, at the $t^{th}$ word.
We introduce a transition score $[A]_{i,j}$ for jumping from $i$ to $j$ tags in successive words, and an initial score $[A]_{i,0}$ for starting from the $i^{th}$ tag.
As the transition scores are going to be trained, we define $\tilde{\theta}=\theta \cup \{ [A]_{i,j} \forall i,j \}$. The score of a sentence $[x]_1^T$ along a path of tags $[i]_1^T$ is then given by the sum of transition scores and networks scores:
\begin{dmath}\label{eq:sentscore}
s([x]^T_1, [i]^T_1, \tilde{\theta}) = \sum_{t=1}^{T}{(A_{[i]_{t-1},[i]_t} + [f_\theta]_{[i]_t,t} )}\,.
\end{dmath}
We normalize this score over all possible tag paths $[j]_1^T$ using a softmax, and we interpret the resulting ratio as a conditional tag path probability.
Taking the log, the conditional probability of the true path $[y]^T_1$ is therefore given by:
\begin{dmath}
 \log{p([y]_1^T, [x]_1^T, \tilde{\theta})} =  s([x]^T_1, [y]^T_1, \tilde{\theta}) - \operatorname*{logadd}_{\forall [j]_1^T} s([x]^T_1, [j]^T_1, \tilde{\theta})\,,
\end{dmath}
where we adopt the notation
\begin{equation}
\operatorname*{logadd}_i z_i = \log{(\sum_i{e^{z_i}})}\,.
\end{equation}
Computing the log-likelihood efficiently is not straightforward, as the number of terms in the  $\operatorname*{logadd}$ grows exponentially with the length of the sentence.
It can be computed in linear time with the Forward algorithm, which derives a recursion similar to the Viterbi algorithm (see \newcite{Rabiner89atutorial}).
We can thus maximize the log-likelihood over all the training pairs $([x]_1^T,[y]_1^T)$ to find, given a sentence $[x]_1^T$, the best tag path which minimizes the sentence score (\ref{eq:sentscore}):
\begin{equation}
\operatorname*{argmax}_{[j]_1^T} s([x]^T_1, [j]^T_1,  \tilde{\theta})\,.
\end{equation}
In contrast to classical CRF, all parameters $\theta$ are trained in a end-to-end manner, by backpropagation through the Forward recursion, following \newcite{collobert:2011b}.

\subsection{Document-level Approach}
\label{doc}

The document-level approach is a document binary classifier, with classes $y \in \{-1,1\}$.
For each document, a set of (trained) filters is applied to the sliding window described in section \ref{sent}.
The maximum value obtained by the $i^{th}$ filter over the whole document is:
\begin{equation}
\operatorname*{max}_t \Big[ w_i[x]^t+b_i\Big]_{i,t}  \;\;\;\;   \mbox{$1 \leq i \leq n_{filter}$}\,.
\end{equation}
It can be seen as a way to measure if the information represented by the filter has been captured in the document or not.
We feed all these intermediate scores to a linear classifier, leading to the following simple model:
\begin{equation}
f_\theta(x)=\boldsymbol{\alpha} \operatorname*{max}_t \Big[W[x]^t+b\Big]\,.
\end{equation}
In the case of movie reviews, the $i^{th}$ filter might capture positive or negative sentiment depending on the sign of $\alpha_i$.
As in section \ref{sent}, we will also consider a non-linear classifier in the experiments.

\paragraph{Training}

The neural network is trained using stochastic gradient ascent.
We denote $\theta$ all the trainable parameters of the network.
Using a training set $\mathcal{T}$, we minimize the following soft margin loss function with respect to $\theta$:
\begin{equation}
\theta \leftarrow \sum_{(x,y)\in \mathcal{T}}{\operatorname*{log}\Big(1 +e^{- yf_\theta(x)}\Big)}\,.
\end{equation}

\subsection{Embedding Fine-Tuning}

As seen in section \ref{method}, the process to compute generic word embedding is quite straightforward.
These embeddings can then be used as features for supervised NLP systems and help to improve the general performance~\cite{TurianRaBe2010,collobert:2011b,Chen2013}.
However, most of these systems cannot tune these embeddings as they are not structurally able to.
By leveraging the deep architecture of our system, we can define a lookup-table layer initialized with existing embeddings as the first layer of the network.

\paragraph{Lookup-Table Layer}
We consider a fixed-sized word dictionary $\mathcal{D}$. Given a sequence of $N$ words ${w_1, w_2, \dots, w_N}$, each word $w_n \in W$ is first embedded into a $d_{wrd}$-dimensional vector space, by applying a lookup-table operation:

\begin{dmath}
LT_W(w_n)=W\left( \begin{array}{lcl} 0, \dots, & 1 &, \dots, 0\\ &\mbox{\small{at index $w_n$}} & \end{array} \right) = {\langle W \rangle}_{w_n}\,,
\end{dmath}
where the matrix $W \in \mathbb{R}^{d_{wrd} \times |\mathcal{D}|}$ represents the embeddings to be tuned in this lookup layer. ${\langle W \rangle}_{w_n} \in  \mathbb{R}^{d_{wrd}}$ is the $w^{th}$ column of $W$ and $d_{wrd}$ is the word vector size. Given any sequence of $N$ words $[w]_1^N$ in $\mathcal{D}$, the lookup table layer applies the same operation for each word in the sequence, producing the following output matrix:

\begin{equation}
LT_W({[w]}_1^N)=\left( {\langle W \rangle}_{[w]_1}^{1} \;\; \dots \;\; {\langle W \rangle}_{[w]_N}^{1} \right)\,.
\end{equation}

\paragraph{Training}
Given a task of interest, a relevant representation of each word is then given by the corresponding lookup table feature vector, which is trained by backpropagation.
Word representations are initialized with existing embeddings.

\section{Experimental Setup}

We evaluate the quality of our embeddings obtained on a large corpora of
unlabeled text by comparing their performance against
the CW \cite{collobert:2008}, Turian \cite{TurianRaBe2010}, HLBL \cite{citeulike:4192404}, and LR-MVL \cite{dhillon11multiviewcca}
embeddings on NER and movie review tasks.
We also show that the general performance can be improved for these tasks by fine-tuning the word embeddings.

\subsection{Building Word Representation over Large Corpora}
\label{wordrep}

Our English corpus is composed of the entire English Wikipedia\footnote{Available at http://download.wikimedia.org. We took the May 2012 version.} (where all MediaWiki markups have been removed), the Reuters corpus and the Wall Street Journal (WSJ) corpus. We consider lower case words to limit the number of words in the vocabulary. Additionally, all occurrences of sequences of numbers within a word are replaced with the string ``NUMBER". The resulting text was tokenized using the Stanford tokenizer\footnote{Available at http://nlp.stanford.edu/software/tokenizer.shtml}. The data set contains about 1,652 million words.
As vocabulary, we considered all the words within our corpus which appear at least one hundred times. This results in a 178,080 words vocabulary.
To build the co-occurence matrix, we used only the 10,000 most frequent words within our vocabulary as context words.
To get embeddings for words, we needed to only consider sequences $T$ of $t=1$ word.
After PCA, each word can be represented in any $n$-dimensional vector (with $n \in \{1,\ldots,10000\}$).
We chose to embed words in a 50-dimensional vector, which is the common dimension among the other embeddings in the literature.
The resulting embeddings will be referred as H-PCA in the following sections.
To highlight the importance of the Hellinger distance, we also computed the PCA of the co-occurence probability matrix with respect to the Euclidean metric. The resulting embeddings are denoted E-PCA.

 \paragraph{Computational cost}
The Hellinger PCA is very fast to compute. We report in Table \ref{tab:comp} the time needed to compute the embeddings described above.
For this benchmark we used Intel i7 3770K 3.5GHz CPUs. As the computation of the covariance matrix is highly parallelizable, we report results with 1, 100 and 500 CPUs. The Eigendecomposition of the $C$ matrix has been computed with the SSYEVR LAPACK subroutine on one CPU. We compare completion times for 1,000 and 10,000 eigenvectors. Finally, we report completion times to generate the emdeddings by linear projection using 50, 100 and 200 eigenvectors. Although the linear projection is already quite fast on only one CPU, this operation can also be computed in parallel.
Those results show that the Hellinger PCA can generate about 200,000 embeddings in about three minutes with a cluster of 100 CPUs.

\begin{table}[h!]
\centering
\begin{tabular}{l | c | c | c}
 & \multicolumn{3}{c}{time (s)}\\
 \hline
 \hline
 \# of CPUs & 1 & 100 & 500 \\
 \hline
 \hline
 Covariance matrix & 9930 & 99 & 20 \\
 \hline
 1,000 Eigenvectors &  72 & - & -\\
 10,000 Eigenvectors & 110 & - & - \\
 \hline
 50D Embeddings & 20 & 0.2 & 0.04 \\
 100D Embeddings & 29 & 0.29 & 0.058 \\
 200D Embeddings & 67 & 0.67 & 0.134\\
 \hline
 \hline
Total for 50D & 10,022 & 171.2 & 92.04 \\
\end{tabular}
\caption{Benchmark of the experiment. Times are reported in seconds.}
\label{tab:comp}
\end{table}

\subsection{Existing Available Word Embeddings}
We compare our H-PCA's embeddings with the following publicly available embeddings:
\begin{itemize}
\item \textbf{LR-MVL\footnote{Available at http://www.cis.upenn.edu/~ungar/eigenwords/}}: it covers 300,000 words with 50 dimensions for each word. They were trained on the RCV1 corpus using the Low Rank Multi-View Learning method. We only used their context oblivious embeddings coming from the eigenfeature dictionary.
\item \textbf{CW\footnote{From SENNA: http://ml.nec-labs.com/senna/}}: it covers 130,000 words with 50 dimensions for each word. They were trained for about two months, over Wikipedia, using a neural network language model approach.
\item \textbf{Turian\footnote{Available at http://metaoptimize.com/projects/wordreprs/}}: it covers 268,810 words with 25, 50, 100 or 200 dimensions for each word. They were trained on the RCV1 corpus using the same system as the CW embeddings but with different parameters. We used only the 50 dimensions.
\item \textbf{HLBL\footnotemark[\value{footnote}] }: it covers 246,122 words with 50 or 100 dimensions for each word. They were trained on the RCV1 corpus using a Hierarchical Log-Bilinear Model. We used only the 50 dimensions.
\end{itemize}

\subsection{Supervised Evaluation Tasks}
\label{tasks}

Using word embeddings as feature proved that it can improve the generalization performance
on several NLP tasks~\cite{TurianRaBe2010,collobert:2011b,Chen2013}.  Using
our word embeddings, we thus trained the sentence-level
architecture described in section \ref{sent} on a NER task.

\paragraph{Named Entity Recognition (NER)} It labels atomic elements in the sentence into categories such as ``PERSON" or ``LOCATION". The CoNLL 2003 setup\footnote{http://www.cnts.ua.ac.be/conll2003/ner/} is a NER benchmark data set based on Reuters data. The contest provides training, validation and testing sets.
The networks are fed with two raw features: word embeddings and a capital letter feature.
The ``caps'' feature tells if each word was in lowercase, was all uppercase, had first letter capital, or had at least one non-initial capital letter.
No other feature has been used to tune the models. This is a main difference with other systems which usually use more features as POS tags, prefixes and suffixes or gazetteers.
Hyper-parameters were tuned on the validation set. We selected $n=2$ context words leading to a window of 5 words. We used a special ``PADDING" word for context at the beginning and the end of each sentence. For the non-linear model, the number of hidden units was 300.
As benchmark system, we report the system of \newcite{Ando05aframework}, which reached 89.31\% F1 with a semi-supervised approach and less specialized features than CoNLL 2003 challengers.

The NER evaluation task is mainly syntactic.
As we wish to evaluate whether our word embeddings can also capture semantic, we trained the document-level architecture described in section \ref{doc} over a movie review task.

\paragraph{IMDB Review Dataset}
We used a collection of 50,000 reviews from IMDB\footnote{Available at http://www.andrew-maas.net/data/sentiment}. It allows no more than 30 reviews per movie. It contains an even number of positive and negative reviews, so randomly guessing yields 50\% accuracy. Only highly polarized reviews have been considered. A negative review has a score $\leq$ 4 out of 10, and a positive review has a score $\geq$ 7 out of 10. It has been evenly divided into training and test sets (25,000 reviews each).
For this task, we only used the word embeddings as features.
We perform a simple cross-validation on the training set to choose the optimal hyper-parameters. The network had a window of 5 words and $n_{filter}=1000$ filters.
As benchmark system, we report the system of \newcite{MaasDPHNP11}, which reached 88.90\% accuracy with a mix of unsupervised and supervised techniques to learn word vectors capturing semantic term-document information, as well as rich sentiment content.

\begin{figure}[h!]
\begin{subfigure}[b]{0.5\textwidth}
\centering
\includegraphics[width=0.8\textwidth]{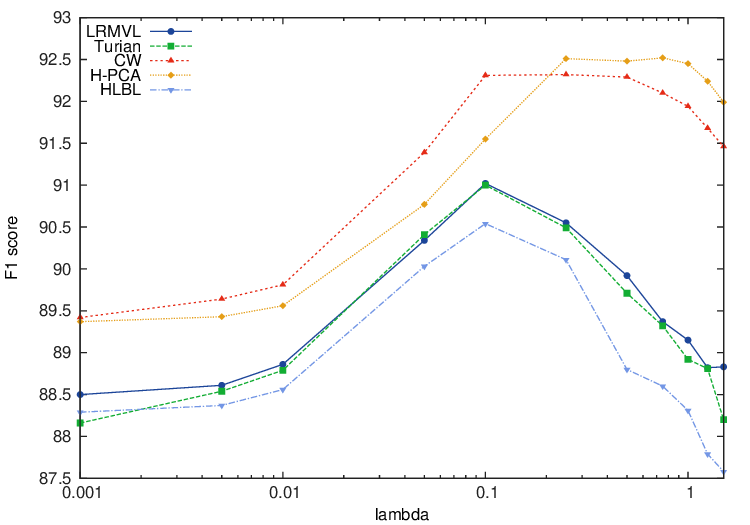}
\caption{NER validation set.}
\end{subfigure}
\begin{subfigure}[b]{0.5\textwidth}
\centering
\includegraphics[width=0.8\textwidth]{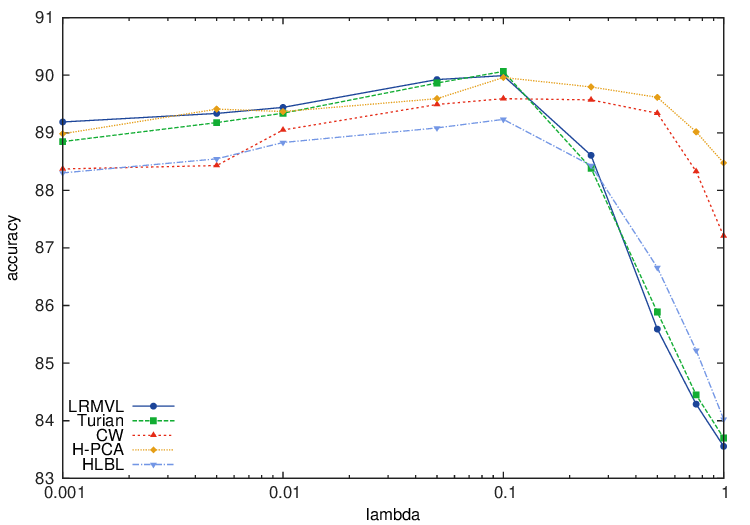}
\caption{IMDB review dataset.}
\end{subfigure}
\caption{Effect of varying the normalization factor $\lambda$ with a non-linear approach and fine-tuning.}
\label{fig:norm}
\end{figure}

\subsection{Embeddings Normalization}

Word embeddings are continuous vector spaces that are not necessarily in a bounded range.
To avoid saturation issues in the network architectures, embeddings need to be properly normalized.
Considering the matrix of word embeddings $E$, the normalized embeddings are:
\begin{equation}
\tilde{E} =\frac{ \lambda (E-\bar{E})}{\sigma(E)}
\end{equation}
where $\bar{E}$ is the mean of the embeddings, $\sigma(E)$ is the standard deviation of the embeddings and $\lambda$ is a normalization factor.
Figure \ref{fig:norm} shows the effect of $\lambda$ on both supervised tasks. The embeddings normalization depends on the type of the network architecture. In the document-level approach, best results are obtained with $\lambda=0.1$ for all embeddings, while a normalization factor set to $1$ is better for H-PCA's embeddings in the sentence-level approach.
These results show the importance of applying the right normalization for word embeddings.

\subsection{Results}
\label{res}
\paragraph{H-PCA's embeddings}
Results summarized in Table \ref{tab:resner} reveal that performance on NER task can be as good with word embeddings from a word co-occurence matrix decomposition as with a neural network language model trained for weeks.
The best F1 scores are indeed obtained using the H-PCA tuned embeddings.
Results for the movie review task in Table \ref{tab:resmovie} show that H-PCA's embeddings also perform as well as all the other embeddings on the movie review task.
It is worth mentioning that on both tasks, H-PCA's embeddings outperform the E-PCA's embeddings, demonstrating the value of the Hellinger distance.
When the embeddings are not tuned, the CW's embeddings slightly outperform the H-PCA's embeddings on NER task.
The performance difference between both fixed embeddings on the movie review task is about 3\%.
Embeddings from the CW neural language model seems to capture more semantic information but we showed that this lack of semantic information can be offset by fine-tuning.

\begin{table}[h!]
\centering
\begin{tabular}{ l | c  | c  }
    \textbf{Approach} & \textbf{Fixed} & \textbf{Tuned} \\
  \hline
  Benchmark & \multicolumn{2}{c }{89.31} \\
  \hline
  & \multicolumn{2}{c }{\textit{Non-Linear Approach}} \\
  H-PCA &  87.91 $\pm$ 0.17 &  89.16 $\pm$ 0.09 \\
  E-PCA &  84.28 $\pm$ 0.15 &  87.09 $\pm$ 0.12 \\
  LR-MVL & 86.83 $\pm$ 0.20  &  87.38 $\pm$ 0.07  \\
  CW & 88.14 $\pm$ 0.21 & 88.69 $\pm$ 0.16 \\
  Turian & 86.26 $\pm$ 0.13 & 87.35 $\pm$ 0.12 \\
  HLBL & 83.87 $\pm$ 0.25 & 85.91 $\pm$ 0.17 \\
  \hline
  \hline
  & \multicolumn{2}{c }{\textit{Linear Approach}} \\
  H-PCA & 84.64 $\pm$ 0.11 & 87.97 $\pm$ 0.09 \\
  E-PCA & 78.15 $\pm$ 0.15  &  85.99 $\pm$ 0.09 \\
  LR-MVL & 82.27 $\pm$ 0.14 & 86.83 $\pm$ 0.17 \\
  CW & 84.50 $\pm$ 0.19 & 86.84 $\pm$ 0.08 \\
  Turian & 83.33 $\pm$ 0.07 &  86.79 $\pm$ 0.11 \\
  HLBL & 80.31$\pm$ 0.11 & 85.06 $\pm$ 0.13 \\
\end{tabular}
\caption{Performance comparison on NER
  task with different embeddings. The first column is results with the original embeddings. The second column is results with embeddings after fine-tuning for this task. Results are reported in
  F1 score (mean $\pm$ standard deviation of ten training runs with different initialization).}
\label{tab:resner}
\end{table}

\begin{table}[h!]
\centering
\begin{tabular}{ l | c  | c  }
    \textbf{Approach} & \textbf{Fixed} & \textbf{Tuned} \\
  \hline
  Benchmark & \multicolumn{2}{c }{88.90} \\
  \hline
  & \multicolumn{2}{c }{\textit{Non-Linear Approach}} \\
  H-PCA &  84.20 $\pm$ 0.16 &  89.89 $\pm$ 0.09 \\
  E-PCA &  74.85 $\pm$ 0.12 & 89.70 $\pm$ 0.06  \\
  LR-MVL & 85.33 $\pm$ 0.14  &  90.06 $\pm$ 0.09  \\
  CW & 87.54 $\pm$ 0.27 & 89.77 $\pm$ 0.05 \\
  Turian & 85.33 $\pm$ 0.10 & 89.99 $\pm$ 0.05 \\
  HLBL & 85.51 $\pm$ 0.14 & 89.58 $\pm$ 0.06 \\
  \hline
  \hline
  & \multicolumn{2}{c }{\textit{Linear Approach}} \\
  H-PCA & 84.11 $\pm$ 0.05 & 89.90 $\pm$ 0.10 \\
  E-PCA & 73.27 $\pm$ 0.16  &  89.62 $\pm$ 0.05 \\
  LR-MVL & 84.37 $\pm$ 0.16 & 89.77 $\pm$ 0.09 \\
  CW & 87.62 $\pm$ 0.24 & 89.92 $\pm$ 0.07 \\
  Turian & 84.44 $\pm$ 0.13 &  89.66 $\pm$ 0.10 \\
  HLBL & 85.34 $\pm$ 0.10 & 89.64 $\pm$ 0.05\\
\end{tabular}
\caption{Performance comparison on movie review task with different embeddings. The first column is results with the original embeddings. The second column is results with embeddings after fine-tuning for this task. Results are reported in classification accuracy (mean $\pm$ standard deviation of ten training runs with different initialization).}
\label{tab:resmovie}
\end{table}

\begin{table*}[ht!]
\centering\sc
\begin{tabular}{ c | c || c | c || c | c  }
\multicolumn{2}{c}{\textbf{boring}} & \multicolumn{2}{c}{\textbf{bad}} & \multicolumn{2}{c}{\textbf{awesome}}\\\
\textit{before} & \textit{after} & \textit{before} & \textit{after} & \textit{before} & \textit{after}  \\
\hline
sad & crap & horrible & terrible & spooky & terrific \\
silly & lame & terrible & stupid & awful & timeless \\
sublime& mess & dreadful & boring & silly & fantastic\\
fancy & stupid & unfortunate & dull & summertime & lovely\\
sober & dull & amazing & crap & nasty & flawless\\
trash & horrible & awful & wrong  & macabre & marvelous\\
loud & rubbish & marvelous & trash & crazy & eerie \\
ridiculous & shame & wonderful & shame & rotten & lively\\
rude & awful & good & kinda & outrageous & fantasy\\
magic & annoying & fantastic & joke & scary & surreal\\
\end{tabular}
\normalfont
\caption{Set of words with their 10 nearest neighbors before and after fine-tuning for the movie review task (using the Euclidean metric in the embedding space). H-PCA's embeddings are used here.}
\label{tab:words}

\end{table*}


\paragraph{Embeddings fine-tuning}
We note that tuning the embeddings by backpropagation increases the general performance on both NER and movie review tasks.
The increase is, in general, higher for the movie review task, which reveals the importance of embedding fine-tuning for NLP tasks with a high semantic component.
We show in Table \ref{tab:words} that the embeddings after fine-tuning give a higher rank to words that are related to the task of interest which is movie-sentiment-based relations in this case.

\paragraph{Linear vs nonlinear model}
We also report results with a linear version of our neural networks.
Having non-linearity helps for NER. It seems important to extract non-linear features for such a task.
However, we note that the linear approach performs as well as the non-linear approach for the movie review task.
Our linear approach captures all the necessary sentiment features to predict whether a review is positive or negative.
It is thus not surprising that a bag-of-words based method can perform well on this task \cite{Wang12}.
However, as our method takes the whole review as input, we can extract \emph{windows} of words having the most discriminative power: it is a major advantage of our method compared to conventional bag-of-words based methods.
We report in Table \ref{tab:wind} some examples of windows of words extracted from the most discriminative filters $\alpha_i$ (positive and negative).
Note that there is about the same number of positive and negative filters after learning.

\begin{table}[ht!]
\centering
\begin{tabular}{ c | p{5cm} }
 $\alpha_i$ &  \multicolumn{1}{c}{$[x]^t$} \\
 \hline
 \hline
 - & \parbox{5cm}{\small \vspace{0.1cm} the worst film this year\\very worst film i 've\\very worst movie i 've \vspace{0.1cm} }\\
 \hline
 - & \parbox{5cm}{\small  \vspace{0.1cm}  watch this unfunny stinker .\\, extremely unfunny drivel come\\, this ludicrous script gets \vspace{0.1cm} }\\
 \hline
 - & \parbox{5cm}{\small  \vspace{0.1cm}  it was pointless and boring\\it is unfunny . unfunny\\film are awful and embarrassing \vspace{0.1cm} }\\
 \hline
 \hline
 + & \parbox{5cm}{\small  \vspace{0.1cm} both really just wonderful .\\. a truly excellent film\\. a really great film \vspace{0.1cm} }\\
 \hline
 + & \parbox{5cm}{\small  \vspace{0.1cm}  excellent film with great performances\\excellent film with a great\\excellent movie with a stellar \vspace{0.1cm} }\\
 \hline
 + & \parbox{5cm}{\small  \vspace{0.1cm}  incredible . just incredible .\\performances and just amazing .\\one was really great . \vspace{0.1cm} }\\
\end{tabular}
\caption{The top 3 positive and negative filters $\alpha_i w_i$ and their respective top 3 windows of words $[x]^t$ within the whole IMDB review dataset.}
\label{tab:wind}

\end{table}

\section{Conclusion}

We have demonstrated that appealing word embeddings can be obtained by computing a \emph{Hellinger PCA} of the word co-occurence matrix.
While a neural network language model can be painful and long to train, we can get a word co-occurence matrix by simply counting words over a large corpus.
The resulting embeddings give similar results on NLP tasks, even from a $N\times10,000$ word co-occurence matrix computed with only one word of context.
It reveals that having a significant, but not too large set of common words, seems sufficient for capturing most of the syntactic and semantic characteristics of words.
As PCA of a $N\times10,000$ matrix is really fast and not memory consuming, our method gives an interesting and practical alternative to neural language models for generating word embeddings.
However, we showed that deep-learning is an interesting framework to fine-tune embeddings over specific NLP tasks.
Our H-PCA's embeddings are available online, here: \texttt{http://www.lebret.ch/words/}.

\section*{Acknowledgments}

This work was supported by the HASLER foundation through the grant ``Information and Communication Technology for a Better World 2020'' (SmartWorld).

\bibliographystyle{acl}
\bibliography{eacl.bll}

\end{document}